%% file: 0-main.tex
\documentclass[conference]{ieeeconf}
\IEEEoverridecommandlockouts

\usepackage[backend=biber,
            hyperref=true,
            url=false,
            isbn=false,
            doi=false,
            backref=true,
            style=ieee,
            citestyle=numeric-comp,
            sorting=none,
            block=none]{biblatex}

\usepackage{flushend}
\usepackage{multirow}
\usepackage{color}
\usepackage{times}
\usepackage{biblatex}

\input{0-preamble.tex}

\usepackage{balance}

\begin{document}
%
\title{Using Intermittent Synchronization to Compensate for Rhythmic Body Motion During Autonomous Surgical Cutting and Debridement}

\author{%
Vatsal Patel*,
Sanjay Krishnan*,
Aimee Goncalves,
Carolyn Chen,\\
Walter Doug Boyd,
Ken Goldberg
\thanks{*Denotes equal contribution. All authors are affiliated with the AUTOLAB at UC Berkeley: \url{autolab.berkeley.edu}.}}

\maketitle

\begin{abstract}
Anatomical structures are rarely static during a surgical procedure due to breathing, heartbeats, and peristaltic movements. 
Inspired by observing an expert surgeon, we propose an intermittent synchronization 
with the extrema of the rhythmic motion (i.e., the lowest velocity windows).
We performed 2 experiments: (1) pattern cutting, and (2) debridement. In (1), we found that the intermittent synchronization approach, while 1.8x slower than tracking motion, was significantly more robust to noise and control latency, and it reduced the max cutting error by 2.6x  In (2), a baseline approach with no synchronization achieves 62\% success rate for each removal, while intermittent synchronization achieves 80\%.

\end{abstract}

\IEEEpeerreviewmaketitle

\input{1-introduction.tex}

\input{2-rw.tex}

\input{5-control.tex}
\input{4-data.tex}

\input{6-experiments.tex}
\input{7-futurework.tex}



{\footnotesize 
\section*{Acknowledgment}
This research was performed at the AUTOLAB at UC Berkeley in
affiliation with the Real-Time Intelligent Secure Execution (RISE) Lab, Berkeley AI Research (BAIR) Lab, and the CITRIS "People and Robots" (CPAR) Initiative: http://robotics.citris-uc.org in affiliation with UC Berkeley's Center for Automation and Learning for Medical Robotics (Cal-MR). This research is supported in part by DHS Award HSHQDC-16-3-00083, NSF CISE Expeditions Award CCF-1139158, by the Scalable Collaborative Human-Robot Learning (SCHooL) Project NSF National Robotics Initiative Award 1734633, and donations from Alibaba, Amazon, Ant Financial, CapitalOne, Ericsson, GE, Google, Huawei, Intel, IBM, Microsoft, Scotiabank, VMware, Siemens, Cisco, Autodesk, Toyota Research, Samsung, Knapp, and Loccioni Inc. We also acknowledge a major equipment grant from Intuitive Surgical and by generous donations from Andy Chou and Susan and Deepak Lim. }

\fontsize{8pt}{9pt}\selectfont
\setlength{\parindent}{0pt}

\begin{flushright}
\printbibliography 
\end{flushright}

\end{document}

%% file: 0-preamble.tex
\usepackage{marginnote}
\usepackage{graphics}
\usepackage[pdftex]{graphicx}
\DeclareGraphicsExtensions{.pdf,.png,.jpg}

\usepackage[rightcaption]{sidecap}
\usepackage{pbox}

\usepackage{mathtools}
\usepackage{amsmath, amssymb, amscd}
\usepackage{ wasysym } 
\usepackage{amsfonts}
\usepackage{mathptmx} 
\usepackage{gensymb} 

\DeclareMathAlphabet{\mathcal}{OMS}{lmsy}{m}{n}
\DeclareSymbolFont{largesymbols}{OMX}{cmex}{m}{n}
\usepackage{textcomp} 

\usepackage[ruled,vlined,linesnumbered]{algorithm2e} 
\usepackage{algorithmicx, algpseudocode} 

\usepackage{array} 
\usepackage{tabularx}
\usepackage{multirow}
\usepackage{multicol}
\usepackage{booktabs}

\usepackage[T1]{fontenc} 
\usepackage[utf8]{inputenc}
\usepackage[english]{babel} 
\usepackage{units}
\usepackage{bm}
\usepackage{times} 
\usepackage{xspace}
\usepackage{flushend}
\usepackage{csquotes}
\usepackage{makeidx}

\let\labelindent\relax
\usepackage{enumitem}

\usepackage{soul} 
\usepackage{subfiles} 

\usepackage[yyyymmdd]{datetime}

\date{\protect\formatdate{1}{1}{2001}}

\usepackage{url}
\makeatletter
\g@addto@macro{\UrlBreaks}{\UrlOrds}
\makeatother
\usepackage{color}
\usepackage[usenames,dvipsnames,table]{xcolor}
\usepackage[pdfborder={0 0 0.5}]{hyperref}
\hypersetup{
    colorlinks=true,
    linkcolor=black,
    citecolor=black,
    filecolor=cyan,
    urlcolor=black
}

\usepackage{soul} 

\newcommand{\todo}[1]{\textcolor{red}{[#1?]}}

\newcommand{\ignore}[1]{}

\newcommand{\seclabel}[1]{\label{sec:#1}}



%% file: 1-introduction.tex
\section{Introduction}
\seclabel{intro}
Robotic Surgical Assistants (RSAs), such as Intuitive Surgical's da Vinci, facilitate precise minimally invasive surgery~\cite{veldkamp2005laparoscopic}. 
RSAs are currently controlled by surgeons using pure tele-operation, often in a master-slave configuration. Operating such robots requires uninterrupted attention, control, and a careful understanding of the robot’s kinematics. 
Automation of surgical sub-tasks has the potential to reduce surgeon tedium and fatigue, operating time, and enable supervised tele-surgery over high-latency networks. 
Several recent papers have proposed techniques for introducing limited autonomy in surgery
almost all in static environments
~\cite{Osa2014, vandenBerg2010, sen2015automating, Schulman2013, swirl2016, murali2015learning, garg2016gpas}.

Virtual simulators are widely used in surgical training to simulate anatomical motions such as breathing, heat beats, or peristaltic movements ~\cite{bielser2004state, steinemann2006hybrid, mendoza2003simulating, sifakis2007arbitrary, nienhuys2001surgery,moglia2016systematic}.
In this paper, we mount the entire surgical workspace on a 6 degrees of freedom miniaturized Stewart platform, allowing the workspace to physically rotate and translate.

Our results suggest that an intermittent synchronization policy, which moves the robot to target around the extrema of the rhythmic motion, is much more robust to kinematic uncertainty and control latency. This is because the minima and maxima of the rhythmic motions correspond to time points with the lowest velocities--thereby small errors in control have lesser effect than at other times.
However, this approach might result in a slower overall execution time.
We used the Stewart platform in an initial pilot study, where an expert cardiac surgeon, W. Doug Boyd, performed a cutting task under rhythmic sinusoidal movement at 0.5 Hz.
We hypothesized that he would attempt to \emph{fully track} the movement of the platform, i.e., mentally model the motion and compensate for it in real time.
However, he preferred an \emph{intermittent synchronization} policy, where he synchronized his actions with the extrema of the rhythmic motion.
This observation was counter-intuitive as intermittent synchronization is less efficient in terms of time than fully tracking the movement of the platform.

\begin{figure}
    \includegraphics[width=\columnwidth]{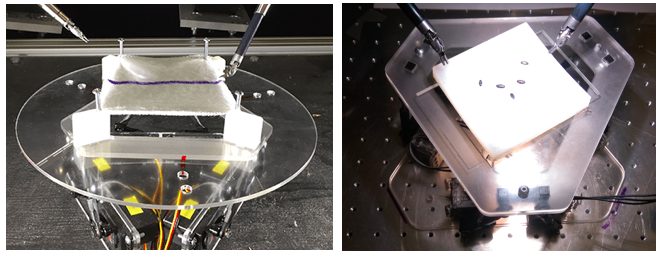}
    \caption{We compare full and intermittent synchronization 
    for two tasks: (1) surgical cutting, and (2) surgical debridement
    on a miniaturized Stewart platform performing rhythmic motions at up to 0.5 Hz with mounted gauze for cutting and black rice seeds on a silicone phantom for debridement. 
    \label{teaser} }
\end{figure}

In this paper, we explore the differences between full and intermittent synchronization in the context of \emph{autonomous} execution on two tasks: (1) surgical cutting, and (2) surgical debridement (Figure \ref{teaser}).
(1) We consider cutting along a line while the platform translates orthogonal to the line sinusoidally at 1 cm, 0.2 Hz.
In our experiments, we constructed a simplified variant of the FLS cutting task, where we autonomously cut along a line and translated the platform perpendicular to the line at 0.2Hz.
The robot had to observe the movement using computer vision, estimate the frequency and phase, and execute a cut along the line.
In our experiments, we found that the intermittent synchronization approach, while 1.8x slower, was significantly more robust.
The maximum cutting error (maximum deviation from the desired trajectory) was reduced by 2.6x.
(2) We consider surgical debridement where foreign inclusions are removed from a tissue phantom that is moving with at 1.25 cm, 0.5 Hz. The robot had to observe the movement, estimate the frequency and phase, and grasp/remove 10 inclusions. A baseline approach achieves 62\% success rate for each removal, while intermittent synchronization achieves 80\%.





%% file: 2-rw.tex
\section{Related Work}
\seclabel{rw}
We use the da Vinci Research Kit (dVRK)~\cite{dVRK,dvrk2014} as our RSA, which is a research platform based on Intuitive Surgical's da Vinci surgical system~\cite{dvrk_firstgen} and which has been frequently used in surgical robotics research~\cite{murali2015learning,vandenBerg2010,sen2016automating,krishnan2017ddco,thananjeyan2017}. 

\subsection{Robot Assisted Surgery with Periodic Motion}
Periodic motion has been studied in robotic surgery including estimation~\cite{ortmaier2005motion, franke2007improved, ortmaier2003motion, richa2010beating, sloth2012model} and control/compensation ~\cite{cavusoglu2005control, duindam2007geometric, riviere2006robotic, moustris2013active}.
Much of this work considers virtual surgical simulators, e.g., Duindam and Sastry \cite{duindam2007geometric}, and proposes a full synchronization approach where the quasi-periodic motion of the anatomy is tracked. Other works, such as Moustris et al. \cite{moustris2013active}, fully synchronize human input on real robot systems with stabilized virtual images, or passively compensate for motion using mounted devices, such as HeartLander \cite{riviere2006robotic}.
Our work considers physical experiments with a physical simulator of periodic motion, which introduces additional challenges of state estimation from imperfect visual signals and control latency. 
We notice that in this setting an intermittent synchronization strategy is more reliable for autonomous execution-- and is consistent with our observations of experts.

\subsection{Experimental Automated Subtasks}
For both of the experimental tasks, cutting and debridement, we adapted state of the art results to new task setting with rythmic motion.
Without periodic motion, surgical robotic cutting has been studied in robotics~\cite{nienhuys2001surgery, murali2015learning, swirl2016,thananjeyanmultilateral, cutting2008} as well as in computer graphics and computational geometry~\cite{zhang2004cutting,Chentanez2009}. 
We constructed a simplified variant of the Fundemental Laparascopic Surgery (FLS) cutting task, where we autonomously cut along a line and translated the platform perpendicular to the line at 0.2Hz.
For this task, without movement, our baseline achieves state-of-the-art cutting accuracy with 100\% success in 40 trials.
The robot had to observe the movement, estimate the frequency and phase, and execute a cut along the line.
This work seeks to extend prior work to a dynamic setting with periodic motion.
Surgical debridement~\cite{Attinger2000_CPMS,Granick2006_WRR,nichols2015} is the process of removing dead or damaged tissue to allow the remaining part to heal. Automated surgical debridement was  demonstrated in~\cite{Kehoe2014, mahler2014case, murali2015learning, daniel2018} without motion.  
As in cutting, our baseline debridement controller with no movement achieves an 85\% success rate--competitive with the state-of-the-art--and we extend this controller to the dynamic setting.

%% file: 5-control.tex
\section{Synchronization Modes}
Next, we formalize the algorithmic problem and present three control strategies.

\subsection{Problem Statement}
We assume a periodic, 1-D sinusoidal, quasi-static movement with negligible higher-order dynamical effects due to motion. The sampling frequency for tracking is assumed to be always greater than $\frac{1}{2T}$, where $T$ is the time period of the rhythmic motion. For full and intermittent synchronization modes, the tracking algorithm uses a colored circle marker.
Let $\mathcal{X} = [g_1,...,g_k]$ denote a sequence of target positions with each $g_i \in SE(3)$ describing a point in a global inertial reference frame.
Each target point is modulated by a discrete time rhythmic motion $m[t] \in SE(3)$ that rotates and translates the all points with respect to the global frame:
\[
g_i^{(m)}[t] = m[t] \times g_i
\]
The period of the motion is defined as the smallest $T$ such that:
\[
\exists T \in \mathbb{N}:  \forall t ~ m[t + T] = m[t],  
\]
and the amplitude is defined as the maximum translational motion due to $m[t]$:
\[
\alpha = \max_{t} \sqrt{m[t].T_x^2  + m[t].T_y^2 + m[t].T_z^2}
\]
We assume the robot has a positional controller and takes decisions of the form $(u, \tau)$ where $u \in SE(3)$ positional control in the global coordinate frame to $\tau$ is a time at which the movement is completed. The robot takes $k$ such decisions $\mathcal{U} = [(u_1, \tau_1),..., (u_k, \tau_k)]$ corresponding to attempted movements to the target points. The error is defined by the cumulative error over all decisions:
\[
\epsilon = \sum_{i=1}^k \| g_i^{(m)}[ \tau ] - u_i \|
\]
Given $\mathcal{X}$ and a position controller, we consider designing a control policy to generate $\mathcal{U}$ when $m[t]$ is unknown.

\subsection{Three Control Modes}
Now, we describe three approaches to synthesizing $\mathcal{U}$.

\subsubsection{No Synchronization (Open Loop)}
The baseline approach is an open-loop control strategy that  ignores motion from the Stewart platform. Mathematically, this is a decision sequence $\mathcal{U}$ that is independent of time:
\[
\mathcal{U} = [(g_1, \cdot), ..., (g_k, \cdot)]
\]
We call this approach no synchronization as it ignores the time at which the the command will terminate.

\subsubsection{Full Synchronization (Full Tracking)}
To translate or rotate the cutting arm to compensate for the motion. The natural first choice of controller is a tracking controller--one that models the motion and tries to exactly compensate for it. 
The next approach tracks the motion of the trajectory and tries to compensate for it.
We consider motion estimation, a two-step process where first the system learns a motion model, then uses that motion model to predict a translational offset. 

First, we track a fixed point $r$ on the platform for an observation period $T_{obs} \gg T$ and a tracking frequency faster than $\frac{1}{2T}$. This involves collecting tuples $(r',r)$, where $r' = m[t] \times r$. 
We perform this tracking with a standard computer vision systems tracks the motion of the Stewart platform at 15 fps for 1 minute. The algorithm segments a colored circle marker with standard OpenCV circle detection on the platform with an endoscopic stereo camera, and register the stereo frame to a pose in a fixed global coordinate frame.

We convert the positions and orientations into $x,y,z$ and Euler angles $y,p,y$ that represent the $SE(3)$ pose.
Then, in each dimension (independently), we fit a 1-D sine curve:
\[C[t] = \alpha \cdot \sin (\omega (t + \phi) ) \]
At any point in time, we correct the robot's commanded motion in each dimension with these offsets:
\[
u_{i}[\tau] = g_i - C[\tau].
\]
This is a 1D approximation to the problem as modeling rotations and translations in the full $SE(3)$ space can be very challenging.
The tracking controller is an efficient solution and has been widely applied in prior work~\cite{cavusoglu2005control, duindam2007geometric, riviere2006robotic, moustris2013active}. However, it is quite hard to implement in a physical setup. This requires precise estimation of $\omega, \phi$ from data. Furthermore, it assumes that commands can be sent with predictable latency to the robot. 

We find that in 10 experimental trials, the estimated frequency has a relative root-mean-square error (RMSE) of $3\%$ in estimating the period due to kinematic variation in the platform and noise in the images. Furthermore, the phase has an RMSE of $0.22$ seconds due to the frame rate of the camera and noise in the images. If we assume that these errors compound at the worst possible time in the trajectory, namely, when $\frac{d}{dt} =1$, then this noise alone could result in $\approx 1$ mm of RMSE. 
Furthermore, we found that the total execution time varied by 0.576 seconds in latency on any particular cut.
This makes it challenging to time the DVRK to a rhythmic motion.
If this error occurred at the worst time ($\frac{d}{dt} =1$), this would result in an additional $> 3$ mm of error.

\subsubsection{Intermittent Synchronization} 
As before, this is a two-step process where first we learn a motion model by fitting a sinusoid to tracking data. 
Then, instead of correcting for the motion at any point of time, as the full synchronization controller, we synchronize the motions with the period of the robot.
We move the robot only so that it reaches the target position around the estimated minima or maxima of the sinusoid.
As before, for each of the 6 degrees of freedom, we fit a sinusoid:
\[C[t] = \alpha \cdot \sin (\omega (t + \phi) ) \]
Then, we identify the ``dominant'' degree of freedom, that is the one with the largest amplitude $\alpha$.
For this degree of freedom, we identify the first maxima or minima and generate a sequence of times based on the inferred at which that optima will occur $[s_1,...,s_k]$ in the future. We synchronize our positional commands with these times:
\[
u_{i}[s_i] = g_i - C[s_i].
\]

While this controller is sub-optimal, in time we hypothesize that it is more robust to estimation and control errors.
For a sinusoid, the change in position is smallest around the optima $\frac{d}{dt} \approx 0$.
In this window, the uncertainty affects the motion the least.
This basic idea actually applies more generally for any smooth continuous disturbance function. For any function, $\frac{d}{dt} \approx 0$ around the optima. The second derivative around the optima ($\frac{d^2}{d^2t}$) can be used to estimate window width.

%% file: 4-data.tex
\begin{figure}[t]
\centering
\includegraphics[width=0.6\columnwidth]{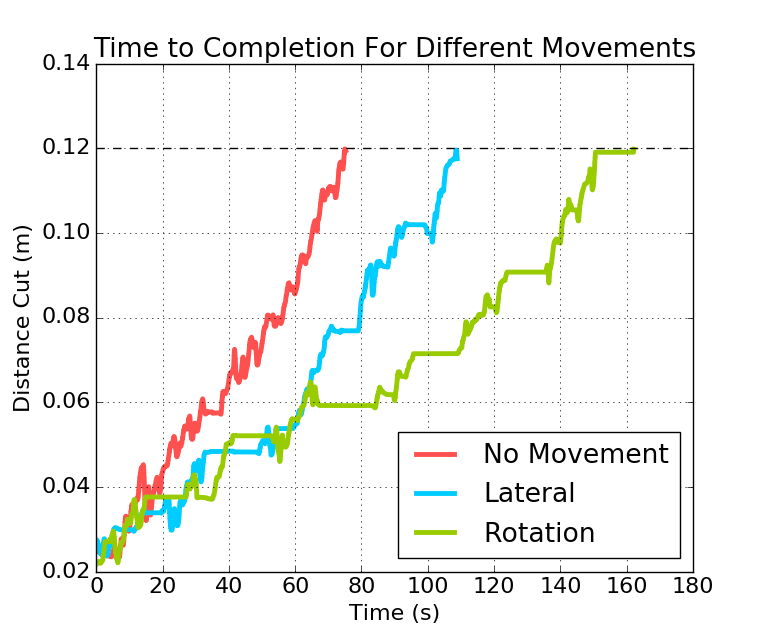}
\caption{Expert human surgeon cutting data from one trial period of experiment 1. Results show the distance along the gauze (12 cm) that was cut over time for the platform with (1a) no motion, (1b) lateral movement, (1c) rotational movement. \label{dataDBoyd} }
\end{figure}

\section{Results: Expert Surgeon Demonstrations}
Observations of a real surgeon motivate the design of the intermittent synchronization controller.
In June 2016, co-author and expert cardiac surgeon Dr. W. Douglas Boyd performed an FLS task (pattern cutting) on the Stewart platform.
We recorded trajectory data from these tasks at 90 measurements per second.
This data consisted of the end effector pose of each of the two DVRK arms, and video from the endoscope.

\subsection{Experiment and Analysis}
In experiment 1, Dr. Boyd cut along a 10 cm line under three types of motion: (1a) no movement, (1b) 0.51 cm, 0.5 Hz sinusoidal lateral movement, and (1c) $10^{o}$, 0.5 Hz rotational movement.
We hypothesized that the surgeon would apply some form of predictive control, and try to compensate for the motion by anticipating where the platform would be.
After running the experiment, we noticed that Dr. Boyd did not track the platform in his control strategy.
Instead, he timed the period of the platform's movement, cutting at the lowest velocity times, i.e, minima and maxima.
This behavior is evident in trajectories with both the rotational and translational motions -- but most pronounced in the rotational motions (Figure \ref{dataDBoyd}).
There was no appreciable reduction in cutting accuracy when Dr. Boyd cut on the moving platform.

Figure \ref{dataDBoyd} plots the the distance cut along the line as a function of time. 
Dr. Boyd completed the baseline task (1a) without any movement in 67 seconds. 
With lateral movement (1b) the task was completed in 139 seconds.
For the rotational movement (1c) the task was completed in 164 seconds.
Waiting for the low-velocity periods results in nearly 3x longer for cutting the entire line with the rotational movements and more than 2x with the translational movements.

%% file: 6-experiments.tex
\begin{table*}[]
\centering
\caption{50 mm line gauze cutting experiments with 25 mm lateral movements at 0.2 Hz. We found that the intermittent synchronization approach, while 1.8x slower, was significantly more robust reducing the max cutting error by 2.6x and was successful in all four trials.}
\label{cut_result}
\begin{tabular}{l||lll||lll||lll||lll}
 & \multicolumn{3}{c}{Trial 1} & \multicolumn{3}{c}{Trial 2} & \multicolumn{3}{c}{Trial 3} & \multicolumn{3}{c}{Trial 4}\\ 
  Controller   & Finish  & Err (mm)         & Time             & Finish   & Err (mm)      & Time              & Finish    & Err (mm)       & Time               & Finish    & Err (mm)         & Time\\ \hline \hline
  No Sync      & No      &  N/A             & 104.51           & No       & N/A           & 100.78            & No        & N/A            & 92.42              & No        & N/A              & 91.54\\
  Full Sync    & Yes     & 5.72             & \textbf{103.35}  & No       & N/A           & 102.21            & No        & N/A            & 97.56              & Yes       & 2.53             & \textbf{96.45} \\
  Int. Sync    & Yes     & \textbf{2.70}    & 206.71           & Yes      & \textbf{1.52} & 181.44            & Yes       & \textbf{1.76}  & 163.35             & Yes       & \textbf{1.12}    & 169.39 \\\end{tabular}
\end{table*}

\section{Experiments: Cutting}
We explore the differences between full and intermittent synchronization in the context of \textbf{autonomous} execution on two tasks: (1) surgical cutting, and (2) surgical debridement. First, we overview our results on (1).

\subsection{Single-Axis Motion}
\label{controlMode}
We illustrate 12 trials of cutting under 2.5 cm, 0.2 Hz sinusoidal motion in one degree of freedom (i.e., single axis).
In the first set of experiments, we evaluated the three control modes for accuracy and reliability.
For each controller, we ran 4 trials of cutting a 5 cm straight line (2 mm thick) in standard surgical gauze.
If during the cutting process, the scissors disengaged from the gauze (either above or below), then the trial was marked as a failure.
We measured the maximum error in cutting as the maximum displacement outside of the 2 mm line.

The no synchronization approach failed in all 4 trials (Table \ref{cut_result}).
The full synchronization approach succeeded 2 out of the 4 times, but incurred a relatively high cutting error.
The intermittent synchronization approach, while 1.8x slower than full synchronization, was successful in all four trials.
It was also significantly more robust and reduced the maximum cutting error by 2.6x.

\subsection{Single-Axis: Increased Frequency}
\begin{figure}[ht!]
\centering
\includegraphics[width=0.6\columnwidth]{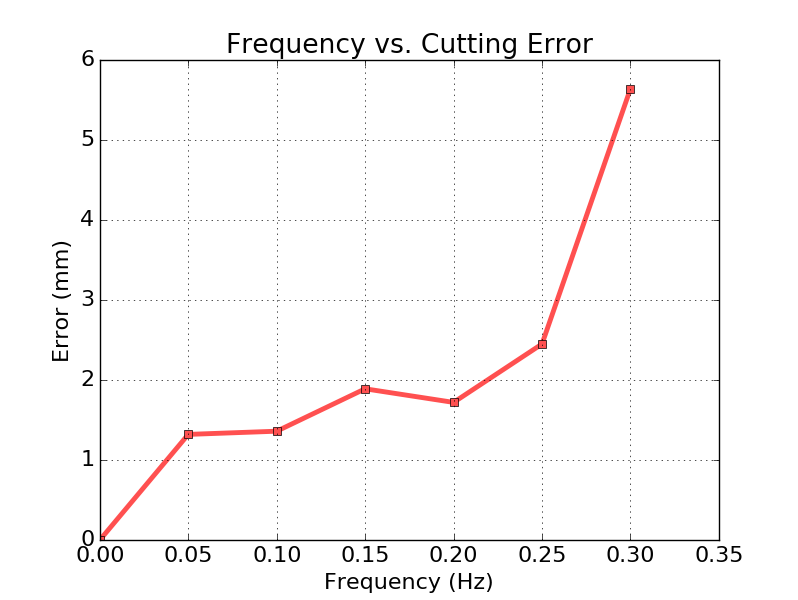}
\caption{ For single-axis motion, the errors are relatively low until a frequency of 0.3 Hz. \label{freq-cut} }
\end{figure}

Next, we performed intermittent frequency cutting experiments over a range of frequencies in platform movement to test for control algorithm uncertainty.
We defined the sinusoidal amplitude to be 0.5 cm and varied the frequency from 0 Hz to 0.3 Hz, which was the highest frequency we could cut at before failure, and measured the error.
We found that for frequencies less than 0.3 Hz the error was relatively low (Figure \ref{freq-cut}, where each data point represents a single trial).
As the frequencies got higher, it became harder to compensate for the motion, and the variability in control latency made it hard to exactly time even the incremental synchronizations.
For example, the error jumps by more than 2x between 0.25 Hz and 0.3 Hz (0.35 Hz fails most of the time).

\subsection{Single-Axis: Alternate Materials}
In the next set of experiments, we considered the same scenario but with alternative cutting materials  to understand how material properties could affect the control technique.
We considered a silicone tissue phantom to model skin and a nylon sheet to model tougher connective tissue.

Over the 10 trials run on the silicone phantoms, 7 trials had an error of 3 mm or less (Table \ref{alt_mat}).
It should be noted that one trial produced an error during cutting and did not finish the experiment.
We attribute the larger cutting deviation errors to frequency estimation errors.

We also repeated the same experiment on 10 nylon sheets (Table \ref{alt_mat}). 
The texture of nylon consists of parallel grains; the cutting trajectories were drawn to go with the grain.
All of the trials were successful and 8 out of 10 trials had an error of 3 mm or less.
As before, we attribute the larger deviations to frequency estimation errors.

Overall, we found that the material properties (gauze vs. silicone vs. nylon) did not affect the error as much as the estimation errors.
In future work, we hope to explore more sophisticated techniques to register the workspace and estimate its movement.


\begin{table}[h]
\centering
\caption{Results from 10 silicone tissue phantom and 10 nylon sheet cutting tests. Out of the 20 cutting trials for the two materials, only one trial failed to complete the task.}
\label{alt_mat}
\begin{tabular}{ccc||ccc}
\multicolumn{3}{c}{\textbf{Silicone Tissue Phantom}}                                                                                                 & \multicolumn{3}{c}{\textbf{Nylon Sheet}}                                                                                                             \\ \hline \hline
\multicolumn{1}{l}{Finish} & \begin{tabular}[c]{@{}c@{}}Trial\\ Err (mm)\end{tabular} & \begin{tabular}[c]{@{}c@{}}Estim\\ Freq (Hz)\end{tabular} & \multicolumn{1}{l}{Finish} & \begin{tabular}[c]{@{}c@{}}Trial\\ Err (mm)\end{tabular} & \begin{tabular}[c]{@{}c@{}}Estim\\ Freq (Hz)\end{tabular} \\ \hline \hline
Yes        & 3                & 12.57               & Yes                  & 2                      & 12.74      \\
Yes        & 6                & 13.42               & Yes                  & 1                      & 13.31      \\
Yes        & 1                & 12.53               & Yes                  & 3                      & 13.40       \\
\textbf{No}  & \textbf{N/A}   & \textbf{13.28}      & Yes                  & 5                      & 12.64       \\
Yes        & 2                & 12.72               & Yes                  & 0                      & 12.79       \\
Yes        & 2                & 12.61               & Yes                  & 4                      & 12.64        \\
Yes        & 1                & 12.66               & Yes                  & 0                      & 12.96       \\
Yes        & 4                & 13.02               & Yes                  & 1                      & 12.59       \\
Yes        & 2                & 19.92               & Yes                  & 2                      & 12.64       \\
Yes        & 2                & 13.08               & Yes                  & 3                      & 12.75                                                    
\end{tabular}
\end{table}

\subsection{Multi-Axis Motions}
We next characterize the performance of intermittent synchronization on multi-axis periodic motions. This means that the periodic motion is in all six degrees of freedom at the same frequency and phase but different amplitudes. 
We compare the following motion modes: No Motion, X only (which translates orthogonally to the cutting line), 3D Translation, and 6D Translation and Rotation. 
For the X only, the motions were sinusoidal with 2.5 cm amplitude and at 0.2 Hz.
For the 3D translation, we generated amplitudes in each of the translation dimensions where the total \emph{norm} of the amplitudes is equal to the amplitude we used for X only. 
For the 6D translation,  we generated amplitudes in each of the translation dimensions where the total \emph{norm} of the amplitudes is equal to the amplitude we used for X only, and the total \emph{norm} of the amplitudes in the rotational dimensions was $15^\circ$. 
We measure the average deviation of a cut from the 2mm marked line over 5 trials for each with the 1 standard deviation error listed below:

\begin{table}[ht!]
\centering
\label{my-label}
\caption{Intermittent Synchronization is best suited for rhythmic translations in a single dominant direction. Performance degrades with periodic motions that rotate and translate in all of SE(3).}
\begin{tabular}{|l|l|}
\hline
        Motion Mode          & Error          \\
\hline
\hline
No Motion              & 0.51  +/- 0.84 \\
X Only                 & 1.66 +/- 1.04  \\
3D Translation     & 3.91 +/- 3.01  \\
6D Trans. and Rot. & 4.41 +/- 2.97 \\
\hline
\end{tabular}
\end{table}

\section{Experiments: Debridement Task}
Next, we consider surgical debridement where foreign inclusions are removed from a tissue phantom that is moving with at 1.25 cm, 0.5 Hz. The motion was along a single axis.
The robot had to observe the movement, estimate the frequency and phase, and grasp/remove 10 inclusions. The inclusions were black rice seeds that were $5$ mm along their longest axis and $2$ mm along the two other axes. We used surgical grippers with a $7$ mm gripper throw. A computer vision system observes the inclusions on a silicone phantom and segments the seeds using a standard contour detector. Then, each seed is registered to the global coordinate frame. Chessboard experiments suggest an inherent error of 2.25mm in the registration system alone.
The robot then controls to the center of mass of the seed and then positions the gripper orthogonal to the long axis. Seeds were placed in random positions and orientations on the surface.

Unsuccessful grasps are unavoidable due to perceptual mistakes and registration/kinematic uncertainty. The task allows retrials and we give the robot a maximum of 20 attempted grasps to clear a phantom of 10 inclusions. We compared using intermittent synchronization to a baseline approach of no synchronization on 10 trials over 10 inclusions. We present the results in the table below.

\begin{table}[h]
\centering
\caption{Results from 10 debridement trials on a baseline no synchronization and 10 debridement trials with intermittent synchronization. }
\begin{tabular}{ccc||ccc}
\multicolumn{3}{c}{\textbf{No Synchronization}}                                                                                                 & \multicolumn{3}{c}{\textbf{Intermittent Synchronization}}                                                                                                             \\ \hline \hline
\multicolumn{1}{l}{Finish} & \begin{tabular}[c]{@{}c@{}}Attempted\\ Grasps\end{tabular} & \begin{tabular}[c]{@{}c@{}}Successful\\ Grasps\end{tabular} & \multicolumn{1}{l}{Finish} & \begin{tabular}[c]{@{}c@{}}Attempted\\ Grasps\end{tabular} & \begin{tabular}[c]{@{}c@{}}Successful\\ Grasps\end{tabular} \\ \hline \hline
No        & 20                & 9               & Yes                  & 16                      &  10     \\
Yes        & 16                &10                & Yes                  & 13                      & 10      \\
Yes        & 10                & 10              & Yes                  & 10                      & 10       \\
Yes  & 15  & 10      & Yes                  & 10                     & 10      \\
No        & 20                & 8               & Yes                  & 13                      & 10      \\
Yes        & 19                & 10               & Yes                  & 10                      & 10       \\
Yes        & 14                & 10               & Yes                  & 11                      & 10       \\
Yes        & 13                & 10               & Yes                  & 10                      & 10      \\
Yes        & 13                & 10               & Yes                  & 10                      & 10       \\
Yes        & 12                & 10               & No                  & 20                      & 9                                                   
\end{tabular}
\end{table}

The baseline approach achieves 62\% success rate for each removal, while intermittent synchronization achieves 80\%. In terms of run time, the intermittent synchronization runs at an average speed of $7.2$ grasps a minute while the baseline runs at $10.1$ grasps a minute.

%% file: 7-futurework.tex
\section{Conclusions}
In this paper, we explored 3 control modes for automated surgical subtasks in a dynamic environment using a Stewart platform, which had a modular design to accommodate different attachments for various testing scenarios.
We analyzed the results of an expert cardiac surgeon cutting gauze on the Stewart platform and extrapolated an intermittent synchronization control strategy, which favored cutting along a trajectory at windows of low velocity. 
In our experiments, we found that the intermittent synchronization approach, while slower, was significantly more robust.